\documentclass[11pt,a4paper]{article}
\usepackage{times,latexsym}
\usepackage{url}
\usepackage[T1]{fontenc}
\usepackage{amsmath, amssymb}
\usepackage{mathrsfs}
\usepackage{multirow}
\usepackage{booktabs}
\usepackage{bbm}
\usepackage{tabularx}
\usepackage{graphicx, float}
\usepackage{xcolor}
\usepackage{colortbl}
\usepackage{bm}
\usepackage{pgfplots}
\usepackage{enumitem}
\usepackage{color}
\usepackage{tikz}
\usepackage[linesnumbered, ruled,vlined]{algorithm2e}
\usepackage{mathabx}

\usepackage[acceptedWithA]{tacl2018v2}
\usepackage{xspace,mfirstuc,tabulary}

\newif\iftaclinstructions
\taclinstructionsfalse 
\iftaclinstructions
\renewcommand{\confidential}{}
\renewcommand{\anonsubtext}{(No author info supplied here, for consistency with
TACL-submission anonymization requirements)}
\newcommand{\instr}
\fi

\iftaclpubformat 

\else

\fi

\title{Neuro-symbolic Natural Logic with Introspective Revision \\for Natural Language Inference}

\author{
Yufei Feng\footnotemark[1] \quad
Xiaoyu Yang\footnotemark[1] \quad
Xiaodan Zhu \quad
Michael Greenspan\\ 
  Ingenuity Labs Research Institute \& ECE, Queen’s University \\ 
 \small{\texttt{\{feng.yufei, xiaoyu.yang, xiaodan.zhu, michael.greenspan\}@queensu.ca}}}
 
\date{}

\begin{document}
\maketitle
\renewcommand{\thefootnote}{\fnsymbol{footnote}} 
\footnotetext[1]{Equal contribution.}
\renewcommand{\thefootnote}{\arabic{footnote}}
\begin{abstract}
We introduce a neuro-symbolic natural logic framework based on reinforcement learning with introspective revision. The model samples and rewards specific reasoning paths through policy gradient, in which the introspective revision algorithm modifies intermediate symbolic reasoning steps to discover reward-earning operations as well as leverages external knowledge to alleviate spurious reasoning and training inefficiency. The framework is supported by properly designed local relation models to avoid input entangling, which helps ensure the interpretability of the proof paths. The proposed model has built-in interpretability and shows superior capability in monotonicity inference, systematic generalization, and interpretability, compared to previous models on the existing datasets. 
\end{abstract}

\section{Introduction}


In the past decade, deep neural networks have achieved impressive performance on modeling natural language inference (NLI)~\cite{dagan05, maccartney2009natural, snli2015, kim2017, esim2017}, which aims to determine the entailment relations between a \textit{premise} sentence and its corresponding \textit{hypothesis}. 
Progress in NLI
has greatly benefited from the models' capabilities at approximating complex underlying functions, discovering and utilizing rich (true and/or spurious) patterns, and exhibiting robustness to noise and ambiguity. However, the black-box models inherently lack interpretability, and still fail to capture many aspects of human reasoning, including monotonicity inference~\cite{help2019, med2019, yanaka2020sys}, systematic compositionality and generalization~\cite{fodor1988connectionism, aydede1997language, yanaka2020sys}, and negation~\cite{geiger2020neural}, among others. 

A recent research trend has attempted to advance the long-standing problem of bringing together the complementary strengths of neural networks and symbolic models \cite{garcez2015neural,yang2017differentiable,rocktaschel2017end,evans2018learning,weber2019nlprolog,de2019neuro,mao2019neuro}. Specifically for natural language, \textit{natural logic}
has long been studied to model reasoning in human language~\cite{lakoff1970linguistics,van1988semantics, valencia1991studies, van1995language, nairn2006computing, maccartney2009natural, maccartney-manning-2009-extended, icard2012inclusion, angeli2014naturalli}.
However, the work of investigating the joint advantage of neural networks and natural logic is sparse \cite{feng2020exploring} (See Sec.~\ref{sec:related_work} for more details) 
and understudied. 

In this paper, we present a neuro-symbolic framework that integrates natural logic with neural networks for natural language inference. 
At the local level, we explore appropriate transformer networks to model the \textit{local} relations between the constituents of a premise and hypothesis, in order to prevent attention from fully entangling the input, which otherwise can seriously impair the interpretability of proof paths built on local relations. 
We then construct natural logic programs and use reinforcement learning to reward the aggregation of the local relations.
When reinforcement learning passes the final reward signals (NLI labels) through the neural natural logic composition network, it faces the challenges of excessive spurious programs (incorrect programs that lead to correct final NLI labels) as well as training inefficiency; the former is particularly harmful to interpretability. Our framework leverages the proposed Introspective Revision method to discover better reward-earning operations and leverage external knowledge to reduce spurious proofs. 

We conducted experiments on six datasets: SNLI~\cite{snli2015}, HELP~\cite{help2019}, MED~\cite{med2019}, MoNLI~\cite{geiger2020neural}, NatLog-2hop~\cite{feng2020exploring}, and a compositional generalization dataset~\cite{yanaka2020sys}. The results show the model’s superior capability in monotonicity inferences, systematic generalization, and interpretability, compared to previous models on these existing datasets, while the model remains a competitive performance on the generic SNLI test set. 

\section{Related Work}
\label{sec:related_work}
\paragraph{Natural Logic:}
Rather than performing deduction over an
abstract logical form, natural logic~\cite{lakoff1970linguistics,van1988semantics, valencia1991studies, van1995language, nairn2006computing, maccartney2009natural, maccartney-manning-2009-extended, icard2012inclusion, angeli2014naturalli} models logical inferences in natural language by operating directly on the structure of
language. Natural logic allows for a wide range of intuitive inferences in a conceptually clean way~\cite{maccartney2009natural, angeli2014naturalli} and hence provides a good framework for developing explainable neural natural language inference models. Specifically, our work is motivated by the natural logic variant proposed by \citet{maccartney-manning-2009-extended}, for which we will provide more background in Sec.~\ref{sec:background}. 

\paragraph{Natural Language Inference:} 
Natural language inference (NLI)~\cite{dagan05, maccartney2009natural, snli2015} aims to identify the entailment relations between the \textit{premise-hypothesis} sentence pairs. 
Benefited from pre-training on large-scale unlabeled corpora and then fine-tuning on large crowd-sourced datasets like SNLI~\cite{snli2015} and MultiNLI~\cite{mnli}, the pre-trained language models~\cite{bert2019, gpt2, radford2018improving} have achieved the state-of-the-art performance.
However, recent work revealed several drawbacks of the current deep NLI systems. The research in \cite{gururangan2018annotation, poliak2018hypothesis} has shown that deep NLI models learn to utilize dataset biases and label-relevant artifacts for prediction. \citet{med2019, help2019, geiger2020neural} showed that a dominating proportion of samples in SNLI and MultiNLI are in upward monotone, and models trained on these datasets have limited abilities to generalize to downward monotone.
More recently, systematically generated datasets have been proposed to evaluate the current models' ability on compositional generalization and showed that pretrained transformers generalize poorly to unseen combinations of the semantic fragments \cite{geiger2019posing, fragments2020,yanaka2020sys, goodwin2020probing}
. 

\paragraph{Neural Network with Logic Components for NLI:} Recent works~\cite{kalouli2020hy, hu2019monalog, chen2021neurallog, feng2020exploring,wu2021weakly} have started to combine neural networks with logic-based components. 
The work most related to ours is \citet{feng2020exploring}, which adapts ESIM~\cite{esim2017} to predict relations between tokens in a premise and hypothesis, and composes them to predict final inferential labels. 
Rather than optimizing the likelihood of specific reasoning paths, the model maximizes the sum of the likelihood of all possible paths (i.e., marginal likelihood) that reach the correct final NLI labels. As a result, the model potentially encourages a large set of \textit{spurious} reasoning paths and has to rely on external prior and strong constraints to predict meaningful intermediate local relations.

This paper, instead, proposes a reinforcement learning with introspective revision framework to sample and reward specific reasoning paths through the policy gradient method. The introspective revision leverages external commonsense knowledge to tackle spurious proof paths and training inefficiency, key issues in developing interpretable neuro-symbolic models. To support that, local relation components need to be carefully designed. We will demonstrate that the proposed model substantially outperforms that proposed in~\cite{feng2020exploring} on five datasets. 

\paragraph{Policy Gradient:} 
Policy gradient algorithms like REINFORCE~\cite{williams1992simple} have been used in neuro-symbolic models to connect neural representation learning and symbolic reasoning~\cite{andreas2017modular,liang-etal-2017-neural,mascharka2018transparency,yi2018neural,mao2019neuro}. 
The original REINFORCE algorithm suffers from sparse rewards and high variances in the gradient. 
To overcome these issues, the research presented in \citet{popov2017data,ijcai2019-331,Trott2019KeepingYD} proposes reward shaping, which leverages domain-specific knowledge to carefully design the reward functions. 
Instead of learning only from the desired outcomes, some approaches also learn from failed attempts. Hindsight Experience Replay (HER)~\cite{NIPS2017_453fadbd} and Scheduled Auxiliary Control (SAC-X)~\cite{riedmiller2018learning} can replay the failed episodes and provide the agent with auxiliary learning goals to enable sample-efficient learning. 
\citet{li2020closed} propose a back-search algorithm, which diagnoses the failed reasoning processes and corrects potential errors to facilitate model training. Based on \citet{li2020closed}, we propose the introspective revision method, which leverages external knowledge to effectively discover reward-earning reasoning programs and to alleviate spurious reasoning.

\section{Background}
\label{sec:background}

\definecolor{lightgray}{gray}{0.90}
\begin{table}
\footnotesize
\centering
\setlength{\tabcolsep}{5pt}
\begin{tabular}{|ccc|}
\hline
\textbf{Relation} & \textbf{Relation Name} & \textbf{Example}  \\
\hline
  $x \equiv y$  & equivalence &  $mom \equiv mother$   \\
  $x \sqsubset y$  & forward entailment &  $cat \sqsubset animal$ \\
  $x \sqsupset y$  & reverse entailment &  $animal \sqsupset cat$  \\
  $x$ \textsuperscript{$\wedge$} $y$  & negation & $human$ \textsuperscript{$\wedge$} $ nonhuman$   \\
  $x \mid y$  & alternation & $cat \mid dog$   \\
  $x \smallsmile y$  & cover & $ animal  \smallsmile nonhuman $  \\
  $x \ \# \ y$  & independence & $ happy \   \#  \ student$   \\
\hline
\end{tabular}
\caption{A set $\mathfrak{B}$ of seven natural logic relations proposed by~\citet{maccartney-manning-2009-extended}.}
\label{tab:seven_relations}
\end{table}

\begin{table}
  \centering
  \setlength{\tabcolsep}{5pt}\footnotesize
\begin{tabular}{|c|c|ccccccc|}

\hline

\rowcolor{lightgray}[5pt][5pt]
\scriptsize{\textbf{Quantifier \&}}&  & \multicolumn{7}{c|}{\textbf{Input Relation $r$}}\\ 

\rowcolor{lightgray}[5pt][5pt]
\scriptsize{\textbf{Connective}} & \multirow{-2}{*}{\textbf{Proj.}}& $\equiv$ &  $\sqsubset$ &   $\sqsupset$ &   $\wedge$  & $\mid$    &   $\smallsmile$  &    $\#$  \\
\hline
\multirow{2}{*}{\textit{all}} & $\rho^{arg1}(r)$ & $\equiv$  &   $\sqsupset$ &   $\sqsubset$ &   $\mid$  & $\#$    &   $\mid$   &    $\#$  \\
 & $\rho^{arg2}(r)$ & $\equiv$  &   $\sqsubset$ &   $\sqsupset$ &   $\mid$  & $\mid$    &   $\#$   &    $\#$  \\
\hline
\multirow{2}{*}{\textit{some}} & $\rho^{arg1}(r)$ & $\equiv$  &   $\sqsubset$ &   $\sqsupset$ &   $\smallsmile$  & $\#$    &   $\smallsmile$   &    $\#$  \\
 & $\rho^{arg2}(r)$ & $\equiv$  &   $\sqsubset$ &   $\sqsupset$ &   $\smallsmile$  & $\#$    &   $\smallsmile$   &    $\#$  \\
\hline
\multirow{2}{*}{\textit{not}} & \multirow{2}{*}{$\rho(r)$} & \multirow{2}{*}{$\equiv$}  &   \multirow{2}{*}{$\sqsupset$} &   \multirow{2}{*}{$\sqsubset$} &   \multirow{2}{*}{$\wedge$}  & \multirow{2}{*}{$\smallsmile$}    &   \multirow{2}{*}{$\mid$}   &    \multirow{2}{*}{$\#$}  \\
 & &    &    &    &    &     &      &      \\
\hline


\end{tabular}
\caption{The projection function $\rho$ maps input relations to output relations under different contexts (here, different surrounding quantifiers).}
\label{tab:projection}
\end{table}

Our model's backbone logic framework
is based on the~\citet{maccartney-manning-2009-extended} variant of the natural logic formalism. The inference system operates by mutating spans of text in a premise to obtain the corresponding hypothesis sentence, and generates proofs based on the natural logic relations of the mutations. 
To extend the entailment relations to consider semantic exclusion, \citet{maccartney-manning-2009-extended} introduced seven set-theoretic relations $\mathfrak{B}$ for modeling entailment relations between two spans of texts (see Table~\ref{tab:seven_relations} for some examples). 

\begin{figure*}[htb]
  \centering
    \includegraphics
    [width=\textwidth,trim={0.3cm 6.8cm 0.3cm 0cm},clip]{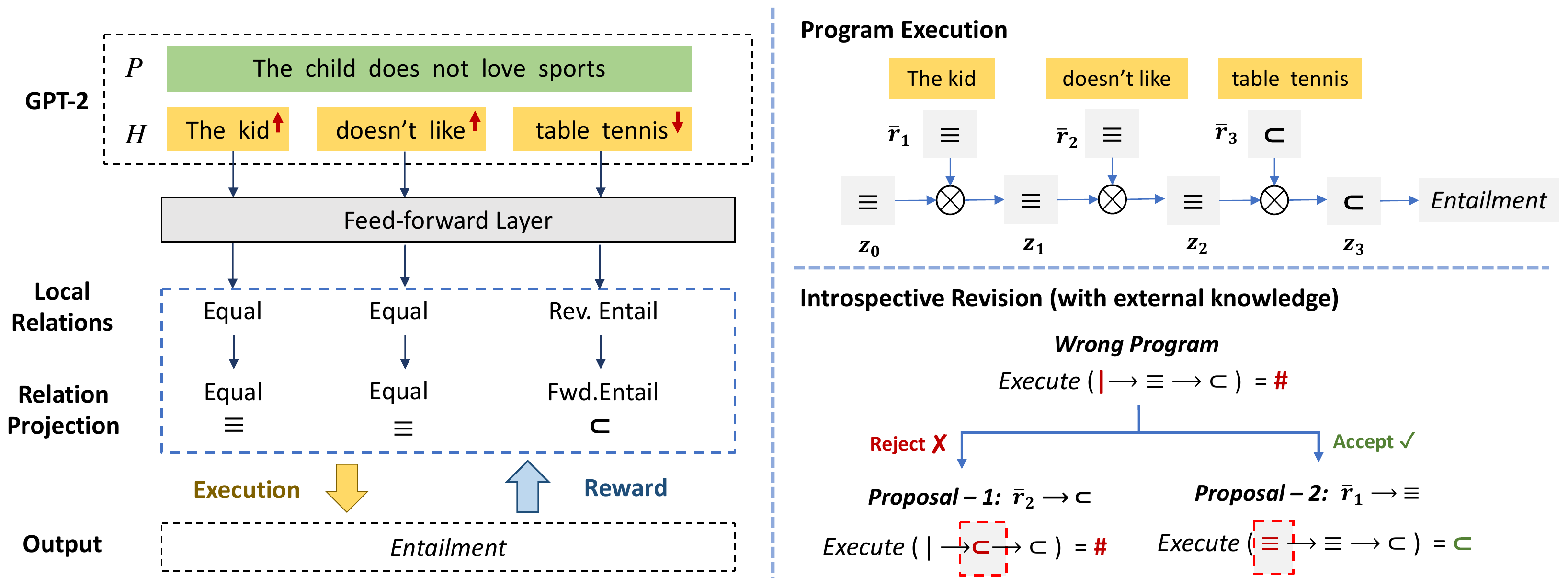}
  \caption{An overview of the proposed neuro-symbolic natural logic framework.} 
\label{main_fig}
\end{figure*}

Assuming the availability of the alignment between a premise and hypothesis, the system first infers the relations between aligned pairs of words or phrases. 
Consider the top-left example in Fig.~\ref{main_fig}: the relation between \textit{``the child''} and \textit{``the kid''} is \textit{equivalence} ($\equiv$), same as the relation between \textit{``does not love''} and \textit{``doesn't like''}, while \textit{``sports''} reversely entails ($\sqsupset$) \textit{``table-tennis''}. 

%

The next step is monotonicity inference. Monotonicity is a pervasive feature of natural language that explains the impact of semantic composition on entailment relations~\cite{van1986essays,valencia1991studies,icard2014recent}. Similar to the monotone functions in calculus, upward monotone keeps the entailment relation when the argument “increases” (e.g., \textit{cat} $\sqsubset$ \textit{animal}). Downward monotone keeps the entailment relation when the argument “decreases” (e.g., in \textit{all animals} $\sqsubset$ \textit{all cats}). The system performs monotonicity inference through a projection function $\rho \colon \mathfrak{B} \rightarrow \mathfrak{B}$, which is determined by the context and projection rules. Table~\ref{tab:projection} shows some examples.
Consider the last row in the table---it shows how the project function $\rho$ works in the negated context following the negation word \textit{not}. Specifically, this row shows seven relations that $\rho(r)$ will output, given the corresponding input relations $r$. For example, if the input relation is \textit{forward entailment} ($\sqsubset$), the function $\rho$ projects it to \textit{reverse entailment} ($\sqsupset$); i.e., $\rho($`$\sqsubset$'$) = $`$\sqsupset$'. As a result, in the example in Fig.~\ref{main_fig}, the \textit{reverse entailment} relation ($\sqsupset$) between \textit{``sports''} and \textit{``table-tennis''} will be projected to \textit{forward entailment} ($\sqsubset$) in the negated context.

\definecolor{lightgray}{gray}{0.90}
\begin{table}
\centering
\small
\begin{tabular}{|>{\columncolor{lightgray}}c|c|c|c|c|c|c|c|}
\hline
\rowcolor{lightgray}
$\Join$ & $\equiv$  &   $\sqsubset$ &   $\sqsupset$ &   $\wedge$  & $\mid$    &   $\smallsmile$   &    $\#$  \\
\hline
$\equiv$ & $\equiv$  &   $\sqsubset$ &   $\sqsupset$ &   $\wedge$  & $\mid$    &   $\smallsmile$   &    $\#$  \\
$\sqsubset$ & $\sqsubset$  &   $\sqsubset$ &   $\#$ &   $\mid$  & $\mid$    &   $\#$   &    $\#$  \\

$\sqsupset$ & $\sqsupset$  &   $\#$ &   $\sqsupset$ &   $\smallsmile$  & $\#$    &   $\smallsmile$   &    $\#$  \\

$\wedge$ &  $\wedge$  &   $\smallsmile$ &   $\mid$ &   $\equiv$  & $\sqsupset$    &   $\sqsubset$   &    $\#$  \\

$\mid$ &  $\mid$  &   $\#$ &   $\mid$ &   $\sqsubset$  & $\#$    &   $\sqsubset$   &    $\#$  \\

$\smallsmile$ & $\smallsmile$  &   $\smallsmile$ &   $\#$ &  $\sqsupset$ &   $\sqsupset$  & $\#$    &    $\#$    \\

$\#$ &   $\#$&   $\#$&   $\#$&   $\#$&   $\#$&   $\#$&   $\#$ \\
\hline
\end{tabular}
\caption{Results~\cite{icard2012inclusion} of composing one relation (row) with another relation (column).}
\label{tab:join}
\end{table}

Built on that, the system aggregates/composes the projected local relations to obtain the inferential relation between a premise and hypothesis. Specifically, Table~\ref{tab:join} shows the composition function when a relation (in a row) is composed with another (in a column). In practice, multiple compositions as such are performed in sequential order or from leaves to root along a constituency parse tree. \citet{maccartney2009natural} shows that different orders of compositions yield consistent results except in some rare artificial cases. Therefore, many works, including ours here, perform a sequential (left-to-right) composition. In the example in Fig.~\ref{main_fig}, composing two \textit{equivalence} ($\equiv$) with \textit{forward entailment} ($\sqsubset$) yields \textit{forward entailment} ($\sqsubset$), resulting in a prediction that the premise entails the hypothesis.

\section{Method}

This section introduces our neural natural logic framework based on the proposed Reinforcement Learning with Introspective Revision approach. We start with local relation modeling, in which caution needs to be taken to avoid the input entangling problem, which can seriously harm the model's interpretability. By viewing the local relation distribution as the stochastic policy, our model then samples and rewards specific reasoning paths through policy gradient, in which the Introspective Revision model can modify intermediate symbolic reasoning steps to discover better reward-earning operations and leverages external knowledge to alleviate spurious reasoning and training inefficiency.

\subsection{Local Relation Modeling}

\label{sec:local_rel}
We use phrases/chunks instead of words as the basic reasoning units. 
The primary motivation
for chunking is to shorten the reasoning paths and hence reduce the number of possible paths, both of which make the reasoning process more efficient. Motivated by \citet{ouyang-mckeown-2019-neural}, we segment the premise $P$ and the hypothesis $H$ into several phrases/chunks. Specifically, we first extract noun phrases with spaCy~\cite{spacy} and then group the continuous spans of words between two noun phrases as chunks. 
As shown in Fig.~\ref{main_fig}, by identifying the noun phrases ``\textit{the kid}'' and ``\textit{table tennis}'', the hypothesis sentence $H$ is segmented into three chunks.
We denote the number of chunks in the hypothesis as $m$, and the $t$-th hypothesis chunk (and its vectorized representation) as $\bm{s}_t$. Similarly the $t'$-th premise phrase is denoted as $\Tilde{\bm{s}}_{t'}$.

As the first step of the neuro-symbolic natural logic, we use a neural network to model the local natural logic relation between each hypothesis phrase $\bm{s}_t$ and its associated premise constituents. However, accurately finding the hard alignment between $\bm{s}_t$ and the corresponding phrase $\Tilde{\bm{s}}_{t'}$ in the premise is a hard problem~\cite{manli}.
Current state-of-the-art NLI systems, like BERT~\cite{bert2019}, use bi-directional soft attention to model the cross-sentence relationship, however, we observe that it tends to fully \textit{entangle} the input~\cite{deyoung2020eraser}.
Consider the top-left example in Fig.~\ref{main_fig}. If we use BERT to encode the input sentences, then the bi-directional attention model can infer the final NLI label solely based on the last-layer hidden states of the first hypothesis phrase ``\textit{the kid}'' because the contextualized representation of this phrase \textit{entangles} the information of the whole input through attention. 
Consequently, the hidden states of the phrase contain global information, thus not being suitable for modeling the local relations.  

To alleviate the undesired entangling, we model local relations with uni-directional attention (such as GPT-2). On the one hand, the uni-directional attention prevents entangling future inputs. 
For example, in Fig.~\ref{main_fig},
the phrase ``\textit{table tennis}'' will not affect the relation prediction anchored on ``\textit{The kid}''. 
On the other hand, although the last hypothesis phrase attends to all previous inputs,
without knowing whether the current phrase is the ending one (the future inputs are not available),
the model cannot skip predicting the natural logic relation at the current phrase $\bm{s}_t$ and postpone all the required reasoning to the last phrase. Specifically, suppose a model always predicts \textit{equivalence} ($\equiv$) at each step $t$ and postpones its final decision to the last hypothesis phrase. Without knowing that ``\textit{table tennis}'' is the ending phrase, the model can predict \textit{equivalence} ($\equiv$) for ``table tennis'' and wait to make a better decision upon seeing the next input phrase, which actually does not exist. 
Failing to make timely local predictions that lead to the correct label before running out of the hypothesis phrases, the model will receive a negative reward in the end.
In this way, the model is encouraged to be more careful in predicting the local relation for each hypothesis phrase. 
We also develop a model that achieves local relations by masking both the past and future hypothesis chunks. Compared to such a model, we will show later (Table~\ref{table:test_acc}) that the uni-directional attention model performs better, partly due to that it preserves the structure of the pretrained GPT-2 model.


Specifically, we propose to model the local relation between $\bm{s}_t$ and the premise $P$, which can be efficiently achieved by the pretrained GPT-2 model~\cite{gpt2}. We concatenate a premise and hypothesis as the input and separate them with a special token $\langle sep \rangle$. The contextualized encoding $\bm{h}_\tau$ for the $\tau$-{th} hypothesis token is extracted from the GPT-2 last-layer hidden states at the corresponding location:
\begin{equation}
    \bm{h}_\tau = GPT\text{-}2(P, H_{1:\tau})
\end{equation}
For the $t$-th phrase in the hypothesis $\bm{s}_t = H_{\tau_1:\tau_2}$, which starts from position $\tau_1$ and ends at position $\tau_2$, we concatenate features of the starting token $\bm{h}_{\tau_1}$ and the ending token $\bm{h}_{\tau_2}$ as the vectorized phrase representation:
\begin{equation}
    \bm{s}_t = Concat(\bm{h}_{\tau_1}, \bm{h}_{\tau_2})
\end{equation}

We use a feed-forward network $f$ with ReLU activation to model the local natural logic relations between the hypothesis phrase $\bm{s}_t$ and its potential counterpart 
in the premise. 
The feed-forward network outputs 7 logits that correspond to the seven natural logic relations listed in Table~\ref{tab:seven_relations}.
The logits are converted with \textit{softmax} to obtain the local relation distribution:
\begin{equation}
    \bm{p}_t = softmax(f(\bm{s}_t)),
\end{equation}
Intuitively, the model learns to align each hypothesis phrase $\bm{s}_t$ with the corresponding premise constituents through attention, and combines information from both sources to model local relations. In practice, the local relation distribution is defined over five relations:
we merge relation \textit{negation} ($\wedge$) and \textit{alternation} ($\mid$) because they have similar behaviors in Table~\ref{tab:join}, and we suppress \textit{cover} ($\smallsmile$), because it is rare in the current NLI datasets. Hence we only need to model five natural logic relation types, following \citet{feng2020exploring}.

\subsection{Natural Logic Program}
\label{sec:nl_program}
We propose to use reinforcement learning to develop neural natural logic, which views the local relation distribution $\bm{p}_t$ as the stochastic policy. At each time step $t$, the model samples a relation $r_t\!\in\!\mathfrak{B}$ according to the policy, and we treat the sequence of sampled relations $\{r_t\}_{t=1}^{m}$ as a symbolic program, which executes to produce the final inferential relation between a premise and hypothesis. According to the best of our knowledge, this is the first model that integrates reinforcement learning with natural logic.

Built on the natural logic formalism of \citet{maccartney-manning-2009-extended}, a projection function $\rho$ (Eq.~\ref{eqn:proj}) maps $r_t$ to a new relation $\bar{r}_t$.
In our model, the projection function $\rho$ is determined by the projectivity feature from the StanfordCoreNLP \textit{natlog} parser\footnote{\url{https://stanfordnlp.github.io/CoreNLP/natlog.html}}. 
For each input token, the projectivity feature specifies the projected relation $\bar{r}_t$ for each input relation $r_t$. In this work, we extend the token-level projectivity to handle phrases: for a phrase with multiple tokens, $\rho$ is determined by the projectivity of the first token in the phrase. In Fig.~\ref{main_fig}, the projectivity of the phrase ``\textit{table tennis}'' is determined by the first token ``\textit{table}'', and $\rho$ projects the predicted \textit{reverse entailment} ($\sqsupset$) relation to \textit{forward entailment} ($\sqsubset$).
\begin{eqnarray}
r_t &=& sampling(\bm{p}_t),\\
\bar{r}_t &=& \rho(r_t) \label{eqn:proj} 
\end{eqnarray}

The program then
composes the projected relations $\{\bar{r}_t\}_{t=1}^m$ to derive the final relation prediction, as shown in top-right part in Fig.~\ref{main_fig}.
Specifically, at time step $t\!=\!0$, the executor starts with the default state $z_0 = equivalence$ ($\equiv$). For each hypothesis phrase $s_{t}, t>0$, the program performs one step update to compose the previous state $z_{t-1}$ with the projected relation $\bar{r}_t$:
\begin{equation}
    z_t = step(z_{t-1}, \bar{r}_t) \label{eqn:execute}
\end{equation}
The final prediction is yielded from the last state $z_m$ of program execution.
Following \citet{angeli2014naturalli}, we group \textit{equivalence} ($\equiv$) and \textit{forward entailment} ($\sqsubset$) as \textbf{\textit{entailment}}; \textit{negation} ($\wedge$) and \textit{alternation} ($|$) as
\textbf{\textit{contradiction}}, and; \textit{reverse entailment} ($\sqsupset$), \textit{cover} ($\cup$), and \textit{independence} ($\#$) as \textbf{\textit{neutral}}.
\paragraph{Rewards and Optimization:}
During training, we reward the model when the program executes to the correct answer. Given a sequence of local relations $\bm{r} = \{r_t\}_{t=1}^m$, at each step $t$ the model receives a reward $R_t$ as follows: 
\begin{equation}
    R_t = 
    \begin{cases}
      \mu, & \text{if}\ Execute(\bm{r}) = y \\
      -\gamma^{m-t}\mu, & \text{if}\ Execute(\bm{r}) \neq y ,\\
    \end{cases}  
    \label{eqn:reward}
\end{equation}
where $\mu$ is the constant reward unit, $\gamma \in (0, 1]$ is the discount factor, and $y$ is the ground-truth label. In addition to Eq.~\ref{eqn:reward}, different rewards are applied under two exceptional cases: (1) if at step $t$ there is no chance for the program to get a positive reward, then the execution is terminated and the model receives an immediate reward $R_t=-\mu$; 
(2) when the true label is \textit{entailment}, the model receives no positive reward if the last state $z_m$ is \textit{equivalence} ($\equiv$). In this way, we encourage the model to select at least one forward entailment ($\sqsubset$) relation during prediction, instead of aggregating a sequence of \textit{equivalence} ($\equiv$) for all entailment cases. In the current NLI datasets, it is less likely that the premise and hypothesis sentences are semantically equivalent to each other.

We apply the REINFORCE~\cite{williams1992simple} algorithm to optimize the model parameters. During training, the local relations $r_t$ are sampled from the predicted distribution, and we minimize the policy gradient objective:
\begin{equation}
    J = - \sum_{t=1}^m \log   \bm{p}_{t}[r_t] \cdot R_t,
    \label{eqn:loss}
\end{equation}
where $\bm{p}_{t}[r_t]$ is the probability that corresponds to the sampled relation $r_t$. During the test, the model picks the relation with largest probability.
\paragraph{Relation Augmentation:}
It can be hard to learn the \textit{reverse entailment} ($\sqsupset$) relation from the existing NLI datasets because the relation of a pair of premise and hypothesis is labeled as \textit{neutral}, if $H$ \textit{entails} $P$ and $P$ \textit{does not entails} $H$.
To help the model distinguish \textit{reverse entailment} $(\sqsupset)$ from \textit{independence} $(\#)$, both of which result in the NLI label \textit{neutral}, we perform relation augmentation to create samples whose hypothesis entails the premise. Specifically, for each sample that is originally labeled as entailment in the training set, we create an augmented sample by exchanging the premise and the hypothesis. 
Note that we avoid augmenting the case where $P$ and $H$ mutually \textit{entail} each other because the new premise still entails the hypothesis after the exchange. To achieve this, we exclude an exchanged sample from relation augmentation if it is still identified as entailment by a pretrained model finetuned on SNLI~\cite{snli2015}.
In terms of the augmented samples, the program receives a positive reward during training if and only if it reaches the correct final state \textit{reverse entailment} $(\sqsupset)$.

\subsection{Introspective Revision}
\label{sec:ir}
The key challenges of developing interpretable neural natural logic models include coping with spurious reasoning paths (incorrect paths $\bm{r} = \{r_t\}_{t=1}^m$ leading to the correct inferential label for a premise-hypothesis pair) as well as training inefficiency.
Finding a correct program that reaches the correct label is challenging because it is inefficient to explore a space of $5^m$ paths for a reward. A positive reward to the correct path is often sparse. 

We propose to use the fail-and-fix approach based on the newly proposed Back-Search algorithm~\cite{li2020closed} to mitigate training inefficiency caused by sparse positive rewards, which, based on a failed program that earns no positive reward, searches for better proof paths in its neighborhood that reaches the correct final prediction. To solve the spurious issue in this fail-and-fix framework, we propose Introspective Revision that leverages external commonsense knowledge (denoted as $\mathcal{K}$) to control spurious proof paths. We believe unstated commonsense knowledge is important not only for improving prediction accuracy (which, as discussed in Sec.~\ref{sec:related_work}, often results from fitting to spurious correlations), but critical for developing interpretable natural language reasoning models by avoiding spurious proofs.

Without loss of generality, we distinguish a non-spurious program $\bm{r}^*$ from spurious ones based on the following assumption, whose effectiveness will be shown and discussed in our experiments. 
\newtheorem{assumption}{Assumption}[section]
\begin{assumption}
A program $\bm{r}^*$ has a larger probability than another program $\bm{r}$ to be a non-spurious program if $\bm{r}^*$ has a better agreement with the external knowledge base $\mathcal{K}$.  
\end{assumption}

\paragraph{External Knowledge:}
 Previous work~\cite{kim2017} queries the knowledge base for each pair of words between a premise and hypothesis exhaustively, which is inefficient and likely to introduce undesired local relations. As a remedy, we found that the lightweight text alignment tool JacanaAligner~\cite{jacana}, though not accurate enough to align all pairs of associated phrases in the input, can be used to guide the search.  
 For a hypothesis phrase $\bm{s}$, we first apply JacanaAligner to obtain its associated premise phrase $\Tilde{\bm{s}}$, and then query the WordNet~\cite{wordnet} database for the possible natural logic relations for the phrase pair $\langle \bm{s}, \Tilde{\bm{s}} \rangle$:
 \newcommand*\circled[1]{\tikz[baseline=(char.base)]{
            \node[shape=circle,draw,inner sep=1pt] (char) {#1};}}
\begin{enumerate}[label=\protect\circled{\arabic*}]
\itemsep0em 
    \item \textbf{Equivalence} ($\equiv$): $\bm{s}$ = $\Tilde{\bm{s}}$ or $\bm{s} \subset \Tilde{\bm{s}}$;
    \item \textbf{Forward Entailment} ($\sqsubset$): $\bm{s} \subset \Tilde{\bm{s}}$ or $\bm{s}$ = $\Tilde{\bm{s}}$ or $\exists\ u \in \bm{s}, v \in \Tilde{\bm{s}}$ and $u$ is a hypernym of $v$;
    \item \textbf{Reverse Entailment} ($\sqsupset$): $\Tilde{\bm{s}} \subset \bm{s}$ or $\exists\ u \in \bm{s}, v \in \Tilde{\bm{s}}$ and $v$ is a hypernym of $u$;
    \item \textbf{Alternation} ($\mid$): $\exists\ u \in \bm{s}, v \in \Tilde{\bm{s}}$ and $u$ is a antonym of $v$;
\end{enumerate}
where $u, v$ denote tokens in the phrase and $\bm{s} \subset \Tilde{\bm{s}}$ means that $\bm{s}$ is a sub-phrase of $\Tilde{\bm{s}}$. The local relations suggested by the knowledge base are formulated as a set of triplet proposals $(t, \Tilde{r}_t, \bm{p}_t[\Tilde{r}])$, where $t$ is the time step, $\Tilde{r}_t$ is the suggested relation, and $\bm{p}_t[\Tilde{r}_t]$ is the model predicted probability that corresponds to $\Tilde{r}_t$. 

Human-curated rules, which are designed to retrieve natural logic relations from the knowledge base, are often imperfect. They inevitably introduce errors due to language variations. For example, intuitively $\bm{s} \subset \Tilde{\bm{s}}$ indicates \textit{forward entailment} ($\sqsubset$); e.g. ``\textit{white cat}'' entails ``\textit{cat}'', while there are cases where the sub-phrase rule indicates \textit{equivalence} ($\equiv$); e.g.,``\textit{have a chat with}'' is equivalent to ``\textit{chat with}'' in meaning. 
In rare cases, the relation can be \textit{alternation} ($\mid$); e.g. ``\textit{fake gun}'' and ``\textit{gun}'' are distinct concepts. While $\bm{s}=\Tilde{\bm{s}}$ often indicates \textit{equivalence} ($\equiv$), our rules need to handle cases where the adverbial is posed in separate phrases; e.g. ``\textit{a bike}'' and ``\textit{near the park}'' entails ``\textit{a bike}''. 
 
To deal with this issue, instead of making an intensive effort to design sophisticated rules to pinpoint a single accurate relation, we design relatively coarse rules to narrow down the possibilities and leave the final choice to the model. Specifically, at each step we provide the model with multiple possible candidates, 
and the proposed introspective revision algorithm introduced in this section decides to accept a useful proposal or reject a misleading one, based on both the reasoning objective (i.e. the label) and the predicted relation distribution.

\paragraph{Algorithm:}
Given a program $\bm{r}\!=\! \{r_t\}_{t=1}^{m}$, the goal of the Introspective Revision algorithm is to find a program $\bm{r}^*$ in the neighbourhood of $\bm{r}$ that executes to the correct answer $y$ while maintaining a large agreement with the external knowledge $\mathcal{K}$, as detailed in Algorithm~\ref{alg:ir}. The algorithm starts with knowledge-driven revision (line 2$\sim$15). We arrange the triplet proposals obtained from the knowledge base as a priority queue $\Phi=\{(t, \Tilde{r}_t, \bm{p}_t[\Tilde{r}_t])\hphantom{0} | \hphantom{0} 0<t\leqslant m,\Tilde{r} \in \mathcal{B} \}$. In each iteration the queue pops the triplet with the largest probability $\bm{p}_t[\Tilde{r}_t]$ that specifies a modification to the sampled program $\bm{r}'\!=\!\bm{Fix}(\bm{r}, t, \Tilde{r}_t)$. In other words, changing the relation $r_t$ at step $t$ of program $\bm{r}$ to the proposed relation $\Tilde{r}_t$ yields a new program $\bm{r}'$. Following~\citet{li2020closed}, the modification is accepted with a probability $1\!-\!\epsilon$ if $\bm{r}'$ executes to the correct answer $y$; otherwise, it is accepted with a probability $\bm{\min}(1, \bm{p}_t[\Tilde{r}_t] / \bm{p}_t[r_t])$. The hyperparameter $\epsilon$ encourages the model to explore low-probability proposals. For each sample, the model accepts or rejects up to $M$ triplets. 

\newcommand\mycommfont[1]{\footnotesize\textcolor{blue}{#1}}
\SetCommentSty{mycommfont}
\SetKwInput{KwInput}{Input}             
\SetKwInput{KwOutput}{Output}      
\SetKw{KwInit}{Init}
\SetKwInput{KwReturn}{Return}  
\SetKwInput{KwParam}{Param}  
\begin{algorithm}[t]
\SetAlgoLined
\small
\DontPrintSemicolon
\KwInput{program $\bm{r} = \{r_1, ... ,r_m\}$,
 label $y$, relation proposals $\Phi$}
\KwParam{maximum step $M$, threshold $\epsilon$}
\KwOutput{$\bm{r}^*$}
\KwInit{$\bm{r}^*= \bm{r}, \hphantom{0}i=0$}\\
\vspace{5pt}
 \tcc{Knowledge-driven revision.}
 \While{$i < M$ and $\Phi \neq \varnothing$}{
  $(t,\hphantom{.}\Tilde{r_t},\hphantom{.}\bm{p}_t[\Tilde{r_t}]) = \Phi.pop()$\\
  sample $u\sim \mathcal{U}(0, 1)$\\
  $\bm{r}' = \bm{Fix}(\bm{r}^*,\hphantom{.}\Tilde{r}_t, \hphantom{.}t)$\\
  \vspace{5pt}
  \tcc{Accept or reject  $(t,\hphantom{.}\Tilde{r_t},\hphantom{.}\bm{p}_t[\Tilde{r_t}])$}
  \eIf{$\bm{Execute}(\bm{r}') = y$ and $u> \epsilon$}{
     $\bm{r}^* = \bm{r}'$\\
   }{
   sample $u\sim \mathcal{U}(0, 1)$\\
   \If{$u < min(1, \hphantom{1}\frac{\bm{p}_t[\Tilde{r}_t]}{\bm{p}_t[r_t]})$}{
   $\bm{r}^* = \bm{r}'$\\
   }
  }
  $i = i + 1$
 }
 \vspace{5pt}
 \tcc{Answer-driven revision.}
 \If{$\bm{Execute}(\bm{r}^*) \neq y$}{
 $\Psi = \bm{GridSearch}(\bm{r}^*, \Phi, y)$\\
 \If{$\Psi \neq \varnothing$}{
 $(t,\hphantom{.}\Tilde{r_t},\hphantom{.}\bm{p}_t[\Tilde{r_t}]) = \Psi.pop()$\\
 $\bm{r}^* = \bm{Fix}(\bm{r}^*, \hphantom{.}\Tilde{r_t}, \hphantom{.}t)$}}
 \KwReturn{$\bm{r}^*$}
 \caption{Introspetive Revision}
 \label{alg:ir}
\end{algorithm}

The knowledge-driven revision above is conservative because only the top-$M$ proposals are considered. However, there are complex cases where the program still cannot reach the correct answer after $M$ steps, or where the provided proposals are insufficient to solve the problem. In these cases, we apply the answer-driven revision (line 17$\sim$22) by conducting a $5\!\!\times\!\!m$ grid search to find modifications that lead to the correct answers. Among the search results $\Psi$, we accept the triplet with the maximum probability. A detailed description of the grid search is presented in Algorithm~\ref{alg:grid}.

Following the reward in Eq.~\ref{eqn:reward} and the objective function in Eq.~\ref{eqn:loss}, we compute a new objective function $J'$ with the modified program $\bm{r}^*$ and its corresponding reward $\bm{R}^*$. 
The model learns by optimizing the hybrid training objective $J_{hybrid}$, defined as:
\begin{eqnarray}
J' = - \sum_{t=1}^m \log   \bm{p}_{t}[r^*_t] \cdot R^*_t\\
    J_{hybrid} = \lambda J + (1-\lambda) J',
\end{eqnarray}
where $\lambda$ is a weight that specifies the importance of the revision. 
The introspective revision algorithm is only applied during training since the label $y$ is required to determine whether a proposal is accepted or not.

\begin{algorithm}[t]
\SetAlgoLined
\small
\DontPrintSemicolon
\KwInput{program $\bm{r} = \{r_1, ... ,r_m\}$,
 label $y$, relation proposals $\Phi$}
\KwOutput{$\Psi$}
\KwInit{$\Psi = PriorityQueue()$}\\
 \For{$t \gets 1$ \textbf{to} $m$}{
 \ForEach{$\Tilde{r} \in \mathfrak{B}$}{
 $\bm{r}'= \bm{Fix}(\bm{r}, \Tilde{r}, t)$\\
 \If{$\bm{Execute}(\bm{r}') = y$}{
 $\Psi.push((t, \Tilde{r}, \bm{p}_t[\Tilde{r}]))$\\
 }
 }
}
\If{$\Psi \cap \Phi \neq \varnothing$}{
$\Psi = \Psi \cap \Phi$}
 \KwReturn{$\Psi$}
 \caption{GridSearch}\label{alg:grid}
\end{algorithm}

\begin{table}[t]
\setlength{\tabcolsep}{5pt}
\scriptsize
\centering

\begin{tabular}{l|l|c}
\toprule
\multirow{2}{*}{\textbf{Phase}} & \multirow{2}{*}{\textbf{Revision (Knowl. / Answ. / Both)}} & \textbf{Success Rate}\\
& & \textbf{of Revision}\\
\midrule
    Start & 80.4\% \phantom{00}( 85.2\% / 8.1\% / \phantom{0}6.7\%) & 59.4\%\\
    End   & 81.7\% \phantom{00}( 80.3\% / 5.9\% / 13.8\%) & 98.4\% \\
\bottomrule
\end{tabular}
\caption{The percentage of samples being revised and the revision success rate at the start/end of the training.}
\label{tab:fix_stat1}
\vspace{5pt}
\setlength{\tabcolsep}{5pt}
\begin{tabular}{l|c|c|c}\toprule
\multirow{2}{*}{\textbf{Relation}}  & \textbf{Knowledge}  & \textbf{Knowl.-driven} & \textbf{Answ.-driven} \\
  & \textbf{Available} & start / end & start / end \\
\midrule
 Equivalence   & 1.035   &   0.595 / 0.482   &  0.096 / 0.026   \\
 Fwd. Entail  &  1.087   &  0.370 / 0.523  & 0.014 / 0.037  \\
 Rev. Entail   & 0.249    &   0.097 / 0.191   &  0.008 / 0.061  \\
 Alternation  &  0.012   &  0.004 / 0.008  & 0.001 / 0.037 \\
 \midrule
 Sum &  2.383   &  1.066 / 1.204  & 0.119 / 0.161
 \\
\bottomrule
\end{tabular}
\caption{The average number of triplet proposals obtained from the WordNet per sample and the average number of proposals accepted by knowledge or answer-driven revision at the start / end of the training.}
\label{tab:fix_stat2}
\end{table}

\section{Experiments}
\begin{table*}
\setlength{\tabcolsep}{7pt}
\small
\centering
\begin{tabular}{l|c|ccccc}\toprule
\textbf{Model}  & \textbf{SNLI}  & \textbf{HELP} & \textbf{MED} & \textbf{MoNLI} & \textbf{Nature Logic-2hop} \\
\midrule
 \textbf{ESIM}~\cite{esim2017}    & 88.0   &   55.3   &   51.8   & 63.9 & 45.1 \\
 \textbf{BERT}-base~\cite{bert2019}   & 90.1   &   51.4   &  45.9   &   53.0 & 49.3\\
  \textbf{GPT-2}~\cite{gpt2} & 89.5   &   52.1   &  44.8   &   57.5 & 48.3\\
   \citet{feng2020exploring}  &      81.2   &   58.2   &  52.4   &   76.8 & 60.1\\
\midrule


\textbf{Ours -- full model} \hfill ~~~\quad\quad\quad (A0)  &  87.5 & \textbf{65.9}   &   \textbf{66.7}  &  \textbf{87.8} & \textbf{62.2} \\

\quad   w/o knowledge \circled{\scriptsize{1}}     &   87.2   &   62.8  &  62.2 & 77.0 & 61.7\\
\quad   w/o knowledge \circled{\scriptsize{2}}    &   87.4   &   65.8  &  64.2 & 81.7 & 51.7 \\
\quad   w/o knowledge \circled{\scriptsize{3}}    &   87.5   &   65.6  &  65.9 & 83.6 & 61.6\\
\quad   w/o knowledge \circled{\scriptsize{4}}     &   87.6   &   65.4  &  64.7 & 83.3 & 58.2\\
\quad  w/o knowledge \circled{\scriptsize{1}}\circled{\scriptsize{2}}\circled{\scriptsize{3}}\circled{\scriptsize{4}}\hfill ~(A1)   &   87.6   &   65.0  &  64.8 & 77.3 & 48.8\\
\quad  w/o answer driven revision \hfill (A2)&  87.5  &  65.4   &   65.5   &  85.1 & 60.9\\ 
\quad  w/o introspective revision \hfill(A3) &  87.6  &  62.1   &   60.7   &   74.4 & 53.3\\ 
\quad  w/o relation augmentation \hfill(A4) &   87.8   &   59.6  &  54.7 & 74.7 & 59.9\\
\midrule
\textbf{Ours} ~~w/ masked attention \hfill ~~(A5)  &  75.9 & 61.3   &   61.6  &  70.9 & 54.6\\
\bottomrule
\end{tabular}
\caption{Model accuracy on multiple challenging test datasets. All models are trained on SNLI and the results of model (A0) $\sim$ (A5) are the average of 3 models starting from different consistently-seeded initializations.}\label{table:test_acc}
\end{table*}

We evaluate the performance of the proposed model on six NLI tasks from various perspectives: the ability of performing monotonicity inference (Sec.~\ref{sec:exp_ood}), reasoning systematicity (Sec.~\ref{sec:exp_sys}), and model interpretability (Sec.~\ref{sec:exp_explain}). 

Our model is trained on Stanford Natural Language Inference (SNLI)~\cite{snli2015}, 
in which the relation between a premise and hypothesis is classified to either \textit{entailment}, \textit{contradiction}, or \textit{neutral}. We set the unit reward $\mu\!\!=\!\!1.0$, and optimize our model with Adam gradient descent for six epochs with a learning rate of \mbox{2e-5}. We compare the models with discount factor $\gamma \in \{0.25, 0.50, 0.75, 1.00\}$ and $\epsilon \in \{0.05, 0.10, 0.20\}$. We found that the test accuracies are not sensitive to $\gamma$ when $\gamma\!\geq\! 0.50$, and we select $\gamma\!\!=\!\!0.50$, $\epsilon=0.20$, which achieved the best validation accuracy on SNLI. For the introspective revision algorithm we set $M\!\!=\!\!3$ based on the average number of proposals (2.383 proposals/sample) in Table~\ref{tab:fix_stat2}. We treat the revised program and the original program as equally informative by setting $\lambda\!\!=\!\!0.5$. Our code is available at \url{https://github.com/feng-yufei/NS-NLI}.

\subsection{Statistics for Introspective Revision}
In Table~\ref{tab:fix_stat1}, we present the statistics for the introspective revision at the start/end of the training, where the natural logic programs are sampled from the predicted distribution. 
Approximately 80\% of the samples perform at least one step of revision, and at the end of the training, there is an increasing chance (98.4\% vs. 59.4\%) that introspective revision helps the model reach the final correct NLI prediction.  
In Table~\ref{tab:fix_stat2}, we show the statistics of the average number of triplet proposals obtained from WordNet and the average number of proposals accepted by knowledge or answer-driven revision during training. \textit{Equivalence} ($\equiv$) and \textit{forward entailment} ($\sqsubset$) make up a large portion of the proposals, while the $alternation$ relation is scarce due to the sparsity of the antonym relation obtained from WordNet. As a result, the numbers of proposals accepted in the knowledge-driven revision are imbalanced across different relations. Moreover, we found that the number of accepted answer-driven revisions slightly increased at the end of the training, which is due to the fact that as the training proceeds, the programs produced by the model are closer to the target labels.  

\subsection{Performance on Monotonicity Reasoning}
\label{sec:exp_ood}
We conduct experiments on multiple recently proposed challenging test datasets for monotonicity inference: HELP~\cite{help2019}, MED~\cite{med2019}, and Monotonicity NLI (MoNLI)~\cite{geiger2020neural}. Unlike SNLI, half of the samples in HELP, MED, and MoNLI are in downward monotone, and they are categorized as \textit{entailment} or \textit{non-entailment}. 
In the above datasets, a premise and the corresponding hypothesis differ by \textit{1-hop}; i.e., they are different by either a lexical substitution, insertion, or deletion. In addition, we also evaluated our model on the Natural Logic 2-hop dataset \cite{feng2020exploring}, which requires a model to perform a 2-hop natural logic composition according to Table~\ref{tab:join}. 

We compare our model with popular natural language inference baselines including ESIM~\cite{esim2017}, BERT-base~\cite{bert2019}, GPT-2\cite{gpt2}, and \cite{feng2020exploring}. 
Following \citet{med2019} and to ensure a fair comparison, all models are trained on SNLI, and during testing, we regard \textit{contradiction} and \textit{neutral} as \textit{non-entailment} if a binary prediction is required.

Table~\ref{table:test_acc} shows the test accuracy on SNLI and four challenging test datasets. Our model performs consistently and significantly better than previous state-of-the-art models on all challenging datasets while achieving competitive ``\textit{in-domain}'' performance on SNLI. Manual inspection shows that compared to GPT-2, a significant proportion of the failure cases on SNLI are due to errors from the projectivity parser, and the ambiguity between \textit{contradiction} and \textit{neutral}~\cite{snli2015}. 
The introspective revision algorithm significantly boosts the model performance on the monotonicity reasoning test sets (A0~vs.~A3). 
Ablation shows that the knowledge-driven revision improves the performance on MoNLI and the 2-hop dataset (A0~vs.~A1), which suggests that without proper constraints, the answer-driven revision can lead to spurious reasoning. We found that removing \textit{equivalence} ($\equiv$) (knowledge \circled{\scriptsize{1}}) from the knowledge-driven revision lowers the performance, because in this case the knowledge-driven revision mistakenly encourages the model to replace \textit{equivalence} ($\equiv$) with \textit{forward entailment} ($\sqsubset$), which may lead to incorrect prediction under downward monotonicity. Compared to \textit{forward entailment} ($\sqsubset$) (knowledge \circled{\scriptsize{2}}), removing \textit{reverse entailment} ($\sqsupset$) (knowledge \circled{\scriptsize{3}}) and \textit{alternation} ($\mid$) (knowledge \circled{\scriptsize{4}}) does not significantly affect the results. We deduce that the relative importance of different relations are affected by the frequency of the external knowledge, and without the help of the knowledge-driven revision, the model can still learn the \textit{reverse entailment} ($\sqsupset$) relation from relation augmentation in Sec.~\ref{sec:nl_program}.
The performance drops when the relation augmentation is vacant (A0 vs. A4).

We also include the model that masks both the past and the future hypothesis chunks in the transformer attention layers for local relation prediction (A5). The model with masked attention yields significantly lower performance on SNLI, partly due to the fact that aggressively masking the past hypothesis chunks changes the structure of the pretrained GPT-2 model, and thus the model benefits less from the pretrained representations.

\subsection{Systematicity of Monotonicity Inference}
\label{sec:exp_sys}
Making systematic generalizations from limited data is an essential property of human language~\cite{lake2018generalization}.
While funetuning pretrained transformers achieves high NLI accuracy, \citet{yanaka2020sys} have recently shown that these models have limited capability of capturing the systematicity of monotonicity inference. We use the dataset proposed by \citet{yanaka2020sys} to evaluate the model's ability in compositional generalization: the model is exposed to all primitive types of quantifiers $\bm{Q}$ and predicate replacements $\bm{R}$, but samples in the training set and test set contain different combinations of quantifiers and predicate replacements. Specifically, with an arbitrarily selected set of quantifiers $\{q\}$ and predicate replacement $\{r\}$, the training set contains data $\bm{D}^{\{q\}, \bm{R}} \cup \bm{D}^{\bm{Q}, \{r\}}$ while the test data only includes the complementary set $\bm{D}^{\bm{Q}\backslash \{q\} , \bm{R}\backslash\{r\}}$. An example of  compositional generalization is shown below:
\begin{enumerate}[label=(\arabic*)]
\itemsep0em 
\small
    \item \textit{P: Some dogs run} $\Rightarrow$ \textit{H: Some animals run}
     \item \textit{P: No animals run} $\Rightarrow$ \textit{H: No dogs runs}
    \item \textit{P: Some small dogs run} $\Rightarrow$ \textit{H: Some dogs run}
\end{enumerate}
An ideal model can learn from the training samples (1), (2), and (3) the entailment relations between concepts \textit{small dog} $\sqsubset $ \textit{dog} $\sqsubset$ \textit{animal}, 
as well as the fact that the quantifier \textit{some} indicates the upward monotonicity and \textit{no} indicates the downward. During testing, the model needs to compose the entailment relations and the monotonicity signatures to make inference over unseen combinations, e.g., sample (4):
\begin{enumerate}[label=(\arabic*)]
\itemsep0em 
\setcounter{enumi}{3}
\small
     \item \textit{P: No dogs run} $\Rightarrow$ \textit{H: No small dogs run}
     \item \textit{P: Near the shore, no dogs run} \hphantom{0}  $\Rightarrow$ \\ \textit{H: Near the shore, no small dogs run}
\end{enumerate}
To test the model stability, \citet{yanaka2020sys} also added adverbs or prepositional phrases as test-only noise to the beginning of both the premise and the hypothesis, e.g., sample (5).

In Table~\ref{table:comp_gen}, all models are trained with 3,270 samples and tested on the complementary test set with about 9,112 examples, exactly following the data split in \citet{yanaka2020sys}. While all baseline models achieved high training accuracy, BERT has limited performance on the test set. For our model, there is only a 3\% gap between the training and test performance, which demonstrates that our model successfully learns to identify and compose the natural logic relations of the predicate replacements with limited training examples. 

We also compare our model to variants of BERT and GPT-2 models that are aware of the token projectivity (models with $\updownarrows$ in Table~\ref{table:comp_gen}). Specifically, for each token, we concatenate the hidden states in the final layer of transformer with its projectivity feature. We aggregated the concatenated features with multiple feed-forward layers and applied average pooling before sending them to the classification layer. Results show that BERT and GPT-2 do not benefit from the projectivity features. The test accuracy drops with additional adverbs and preposition phrases, leaving space for future research on the robustness to unseen perturbations.

\begin{table}
\small
\centering
\begin{tabular}{l|c|ccc}\toprule
\textbf{Model}  & \textbf{Train}  & \textbf{Test} & \textbf{Test$_{adv}$} & \textbf{Test$_{pp}$}\\
\midrule
 \textbf{BERT}-base   & 100.0   &   69.2   &  50.8 & 49.3  \\
 \textbf{GPT-2}  &  100.0   &  25.6  & 35.6  & 35.4 \\
 \midrule
 \textbf{BERT}-base$^{\updownarrows}$   & 100.0    &   65.4   &  51.4 & 52.7  \\
 \textbf{GPT-2}$^{\updownarrows}$  &  100.0   &  28.1  & 35.1 &  39.6\\
 \midrule
 \textbf{Ours} w/o. IR &      91.3   &  79.3 & 57.1     & 54.0  \\
 \textbf{Ours}  &      98.4   &   95.1   &  61.0 & 61.5 \\
\bottomrule
\end{tabular}
\caption{Results for compositional generalization; ${\updownarrows}$ marks the models with polarity features. }\label{table:comp_gen}
\end{table}

\subsection{Evaluation of Model Explainability}
\label{sec:exp_explain}
\begin{table*}
\small
\centering
\begin{tabular}{l|ccc|c}\toprule
 \multirow{2}{*}{\textbf{Model}} & \textbf{e-SNLI} & \textbf{e-SNLI} & \textbf{e-SNLI} &  \textbf{Natural Logic 2-hop} \\
  & IOU  & Precision / Recall / F1 & Human Eval. & Acc.\\
\midrule
 \citet{lei2016rationalizing}   & 0.42   &   0.37 / 0.46 / 0.41   & 56 / 100 & --    \\
 \citet{feng2020exploring}  &  0.27   &  0.21 / 0.35 / 0.26  & 52 / 100 & 0.44  \\
 \midrule
 \textbf{Ours} -- full model \hfill (B0) &      \textbf{0.44}   &   \textbf{0.58} / \textbf{0.49} / \textbf{0.53}  &  \textbf{80} / 100 & \textbf{0.52}  \\ 
 \quad w/o. external knowledge \hfill (B1)& 0.41 & 0.53 / 0.45 / 0.48 & 67 / 100 &  0.44 \\
 \quad w/o. introspective revision \hfill (B2)& 0.40 & 0.52 / 0.43 / 0.47 & 68 / 100 & 0.43 \\
 \quad w/o. relation augmentation \hfill (B3) & \textbf{0.44}& 0.57 / 0.48 / 0.52 & 75 / 100 & 0.51 \\
 \textbf{Ours} -- BERT encoder \hfill (B4) &      0.14   &   0.20 / 0.15 / 0.17  &  29 / 100 & 0.28  \\ 
\bottomrule
\end{tabular}
\caption{Evaluation for the model generated explanation.}\label{table:explain}
\end{table*}


The proposed model provides built-in interpretability following natural logic---the execution of programs $\{z_t\}_{t=1}^m$ (Eq.~\ref{eqn:execute}) provides explanation along with the model's decision making process, namely giving a \textit{faithful} explanation~\cite{jacovi-goldberg-2020-towards}.
To evaluate the model interpretability, we derive the predicted rationales from the natural logic programs and compare it
with human annotations in e-SNLI~\cite{esnli}. Specifically, our model regards as rationales the hypothesis phrases $\bm{s}_t$ that satisfies: (1) $z_t$ points to the final prediction according to the grouping described at the end of Sec.~\ref{sec:nl_program}; (2) $z_t \neq z_{t-1}$. Following \citet{deyoung2020eraser}, we use Intersection Over Union (IOU) formulated in Eq.~\ref{eqn:global_iou} as the evaluation metric: the numerator is the number of shared tokens between the model generated rationales and the gold rationales, and the denominator is the number of tokens in the union. We also compute finer-grained statistics over individual rationale phrases. Following \citet{deyoung2020eraser}, a predicted rationale phrase $p$ matches an annotated rationale phrase $q$ when $IOU(p, q) \geqslant 0.5$, and we use \textit{precision}, \textit{recall} and \textit{F1 score} to measure the phrasal agreement between the predicted rationales and human annotations. We also invited 3 graduate students (not the authors of this paper) to evaluate the quality of the predicted rationales on the first 100 test samples in e-SNLI. Given the premise-hypothesis pair and the golden label, the evaluators judged the explanation as plausible if the predicted rationale (1) alone is sufficient to justify the label, and; (2) does not include the whole hypothesis sentence.  
\begin{equation}
\small
    IOU = \frac{num\text{-}tokens \{ R_{pred} \cap R_{truth}\}}{num\text{-}tokens \{ R_{pred} \cup R_{truth}\}}
    \label{eqn:global_iou}
\end{equation}

From the perspective of natural logic, we follow \cite{feng2020exploring} to evaluate the quality of the natural logic programs.
For each sample, the Natural Logic 2-hop dataset provides the gold program execution states, and we evaluated the accuracy of our predicted states $\hat{z}_t$ against the ground-truth.
We compare our model with representative neural rationalization models proposed by~\citet{lei2016rationalizing}, which learns to extract rationales without direct supervision, and \citet{feng2020exploring}, which explains its prediction by generating natural logic reasoning paths. 
The summary statistics in Table~\ref{table:explain} shows that our model matches \citet{lei2016rationalizing} on the IOU score, and that it produces rationales with significantly higher precision and F1-scores on the e-SNLI test set. The superior rationalization performance is also supported by the human evaluation mentioned above (the $4$th column in Table~\ref{table:explain}).
Compared to \citet{feng2020exploring}, our model produces intermediate natural logic states that better agree with the ground truth.
The results in Table~\ref{table:explain} show that the model explanation significantly benefits from the external knowledge (B0 vs. B1), and the answer-driven revision alone does not improve the quality of the generated rationales (B1 vs. B2).
We also compare our model to the system that replaces the uni-directional attention model GPT-2 with the bi-directional attention model BERT. The model with BERT encoder yields significantly lower scores on interpretability (B0 vs. B4).

\begin{figure*}[htb]
  \centering
    \includegraphics
    [width=\linewidth,trim={0cm 4.8cm 5.9cm 0cm},clip]{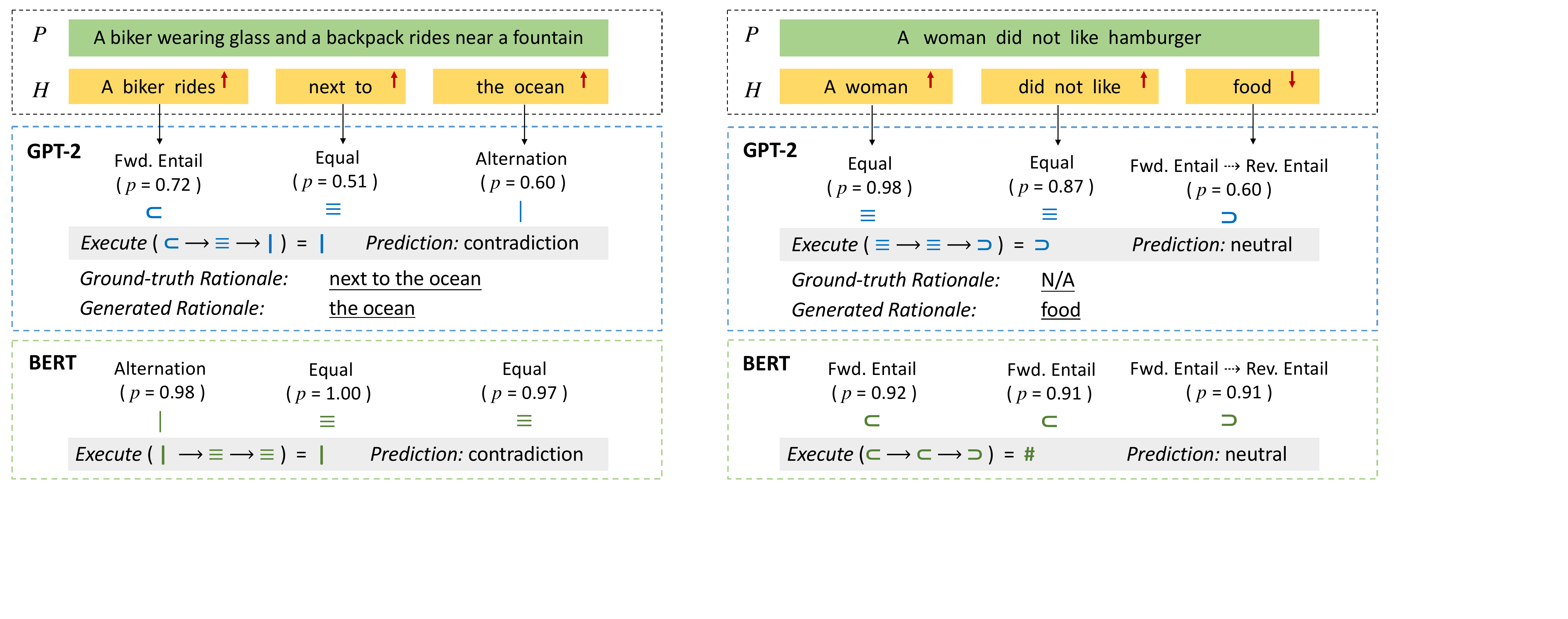}
  \caption{Examples for predictions and explanation for some cases from SNLI (left) and MoNLI (right).} 
\label{fig:case}
\end{figure*}

\subsection{Case Study}
\label{sec:case}
The upper part of the Fig.~\ref{fig:case} shows how our natural logic model makes predictions during testing.
The left example involves upward monotone. Upon seeing the premise and the first hypothesis phrase \textit{A biker rides}, the model predicts the local relation as \textit{forward entailment} ($r_1 = $`$\sqsubset$') at time step $t\!\!=\!\!1$. The predicted relation stays unchanged after applying the projection function $\rho($`$\sqsubset$'$) = $`$\sqsubset$' because it is in the context of upward monotone. According to Table~\ref{tab:join} we have $z_1\!\!=\!\!z_0 \otimes r_1\!\!= $`$\sqsubset$'. Similarly, as the second prediction for the phrase \textit{next to}, relation \textit{equivalence} ($r_2 = $`$\equiv$') does not change the reasoning states because $z_2\!\!=\!\!z_1 \otimes r_2\!\!= $`$\sqsubset$'. The third hypothesis phrase \textit{the ocean} is a distinct concept against \textit{a fountain} in the premise, our model outputs relation \textit{alternation} ($r_3 = $` $ \mid$ ') and we have $z_3 = z_2 \otimes r_3 = $` $\mid$ '. The model runs out of the hypothesis phrases after 3 steps, and reaches \textit{contradiction} according to the final state $z_3$. 

An additional example with downward monotone 
is illustrated on the right of Fig.~\ref{fig:case}. Our model predicts the relation \textit{forward entailment} ($r_3 = $`$\sqsubset$') at the third time step since \textit{food} includes \textit{hamburger}. 
The projection function flips the relation to \textit{reverse entailment} ($\sqsupset$) because according to the projectivity
in Table~\ref{tab:projection}, the first argument that follows negation \textit{did not} is in downward monotone, i.e., $\rho($`$\sqsubset$'$) = $`$\sqsupset$'.

At the bottom of Fig.~\ref{fig:case}, we provide examples for the reasoning processes produced by the natural logic model that is built upon the bi-directional attention model BERT. 
Although it produces the same final labels as our proposed model, the model based on BERT can predict wrong local relations due to its entangling effect.
Specifically, the model with bi-directional attention is prone to make its final decision in the first place (82\% of the cases in the human evaluation), and then predict local relations that can keep the initial decision during the program execution (according to the composition rules in Table~\ref{tab:join}). In the first example in Fig.~\ref{fig:case}, to keep the first predicted relation \textit{alternation} ($\mid$) unchanged during execution, the model subsequently predicts a series of \textit{equivalence} ($\equiv$) relations. In the second example, the model predicts local relation \textit{forward entailment} ($\sqsubset$) for each hypothesis phrase, and at the last step, the \textit{forward entailment} ($\sqsubset$) relation is projected to \textit{reverse entailment} ($\sqsupset$) according to the projectivity.

\section{Summary}
The proposed neuro-symbolic framework integrates the long-studied natural logic with reinforcement learning and introspective revision, effectively rewarding the intermediate proof paths and leveraging external knowledge to alleviate spurious reasoning. The model has built-in interpretability following natural logic, which allows for a wide range of intuitive inferences easily understandable by humans. Experimental results show the model's superior capability in monotonicity-based inferences and systematic generalization, compared to previous models on the existing datasets, while the model keeps competitive performance on the generic SNLI test set. 

\section*{Acknowledgements}
This research was supported by NSERC Discovery Grants. We thank the anonymous reviewers and action editors for their helpful comments.

\bibliography{tacl2018}
\bibliographystyle{acl_natbib}

\end{document}

%